\documentclass[sigconf]{acmart}
\AtBeginDocument{%
  }

\setcopyright{acmlicensed}
\copyrightyear{2018}
\acmYear{2018}
\acmDOI{XXXXXXX.XXXXXXX}
\acmConference[Conference acronym 'XX]{Make sure to enter the correct
  conference title from your rights confirmation email}{June 03--05,
  2018}{Woodstock, NY}
\acmISBN{978-1-4503-XXXX-X/2018/06}

\usepackage{algorithm}
\usepackage{amsmath}
\usepackage{tcolorbox}
\usepackage{subcaption}
\usepackage{booktabs}
\usepackage{multirow}

\begin{document}

\title{Do LLMs Know Your Neighborhood? Auditing LLM Priors for Neighborhood-Level Mobility Prediction and Structural Alignment}

\author{Saad Mohammad Abrar}
\email{sabrar@terpmail.umd.edu}
\orcid{0000-0002-3037-0016}
\affiliation{%
  \institution{University of Maryland}
  \country{USA}
}

\author{Eesha Kurella}
\email{ekurella@umd.edu}
\affiliation{%
  \institution{University of Maryland}
  \country{USA}
}

\author{Arnav Dadarya}
\email{arnav1@terpmail.umd.edu}
\orcid{0000-0001-6036-076X}
\affiliation{%
  \institution{University of Maryland}
  \country{USA}
}

\author{Naman Awasthi}
\email{nawasthi@terpmail.umd.edu}
\orcid{0000-0001-6036-076X}
\affiliation{%
  \institution{University of Maryland}
  \country{USA}
}

\author{Kazi Tasnim Zinat}
\email{kzintas@terpmail.umd.edu}
\orcid{0000-0001-6036-076X}
\affiliation{%
  \institution{University of Maryland}
  \country{USA}
}

\author{Vanessa Frias-Martinez}
\email{vfrias@umd.edu}
\orcid{0000-0001-5114-7633}
\affiliation{%
  \institution{University of Maryland}
  \country{USA}
}
\renewcommand{\shortauthors}{Abrar et al.}

\begin{abstract}
  Human mobility is central to urban planning, transportation management, public health, and emergency response, yet the fine-grained trajectory data needed to model movement are often proprietary, access-restricted, and privacy-sensitive, motivating the search for alternatives. Recent work suggests that large language models (LLMs) offer one such alternative, generating plausible mobility traces and predicting individual movement. However, little is known about whether they can infer aggregate mobility patterns across neighborhoods, which reveal collective dynamics that can directly inform public health response, transportation modeling, and emergency planning. Beyond inference, whether these predictions capture empirically meaningful context-mobility relationships also remains open. We address this gap by evaluating zero-shot LLMs on Census Block Group-level mobility prediction across four U.S. metropolitan areas, using anonymized Cuebiq mobility traces to construct point-level, trajectory-level, and temporal mobility outcomes and pairing them with sociodemographic and built-environment predictors. We compare LLM predictions against supervised baselines and introduce a directional alignment analysis that tests whether LLM-implied predictor effects agree with empirical OLS and Jonckheere–Terpstra trends. Results show that neighborhood context contains substantial predictive signal, with supervised baselines reaching 0.580 average accuracy compared with 0.435 for the best LLM, while spatial extent outcomes are the most predictable but also exhibit the largest LLM–baseline gaps. Directional analysis further shows that LLMs rely on coarse, stable predictor-level priors that often remain invariant across outcomes and cities, including asymmetric treatment of protected-group predictors. These findings suggest that LLMs can partially recover aggregate mobility patterns from urban context, but their predictions should not be treated as structurally grounded without auditing their empirical alignment and potential bias.

\end{abstract}

\ccsdesc[500]{Information systems~Spatial-temporal systems}
\ccsdesc[300]{Computing methodologies~Supervised learning}
\ccsdesc[300]{Information systems~Data mining}

\keywords{Large language models, human mobility, neighborhood-level mobility, urban computing, geospatial AI, directional alignment, mobility prediction, algorithmic auditing}


\maketitle

\section{Introduction}
\label{sec:intro}
Human mobility is a fundamental dimension of urban life, reflecting how people access opportunities, navigate infrastructure, and experience cities in everyday life \cite{barbosa2018human, hagerstrand1970people, yang2023identifying, du2025review}. Because movement patterns shape access to jobs and services, pandemic spread, environmental burdens, and broader forms of social interaction, mobility data have become central to understanding cities and informing urban policy \cite{abrar2023analysis, buckee2020aggregated, chang2021mobility, yabe2025behaviour, xu2025using}. The growing availability of passively collected digital traces, especially smartphone- and mobile-phone-based location data, has substantially expanded this potential. Compared with traditional travel surveys \cite{bricka2024summary} and other coarse data sources, such data provide much finer spatial and temporal resolution, often capturing behavior continuously over longer periods and at much larger scales \cite{gonzalez2024using, garber2022selection}. This has enabled researchers to characterize everyday activity patterns, travel extent, accessibility, and behavioral heterogeneity with unprecedented detail, greatly enlarging the empirical toolkit available to urban science and policy research.

Yet the smartphone-based location data that enable these analyses are largely proprietary, access-restricted, and subject to privacy constraints that limit their availability to many researchers and public agencies, raising the question of whether models that encode broad knowledge about cities and human behavior could serve as useful alternatives in data-scarce settings. Large language models (LLMs) are a natural candidate for this role. Trained on vast corpora encompassing urban descriptions, census and demographic data, transportation research, and accounts of everyday life across cities, LLMs may encode the kinds of contextual knowledge that underlie mobility patterns. Recent studies suggest they do. Prior work has used LLMs for tasks such as next-location prediction and zero-shot mobility inference, showing that LLMs can sometimes act as mobility predictors even without task-specific training \cite{wang2023would, beneduce2025large, ma2025learning}. Other work has explored whether prompted LLMs can generate synthetic travel-survey responses for urban mobility assessment, producing plausible diary-like behavior from background knowledge alone \cite{bhandari2024urban}. More broadly, benchmarks such as CityBench evaluate LLMs across a range of urban tasks and show that, while advanced models can be competitive on tasks grounded in commonsense and semantic understanding, they remain less reliable on tasks requiring stronger urban reasoning \cite{feng2025citybench}. However, all of these studies share a common target of evaluation, assessing whether LLMs can produce plausible outputs about individual-level mobility, such as a person's next trip, a synthetic travel diary, or a single device's behavioral signature.

This individual-level focus is valuable in its own right, but it does not fully address the forecasting needs of many real-world systems, where the key quantity of interest is aggregate mobility patterns. Transportation agencies, urban planners, emergency managers, and public health officials typically require predictions of origin-destination flows, demand surges, and population redistribution patterns rather than individual trajectories \cite{zhao2024origin,rong2024interdisciplinary}. Although one possible approach is to simulate many individual LLM-based agents and aggregate their generated trajectories, this strategy can be computationally expensive. We instead study whether LLMs can directly infer aggregate mobility outcomes from contextual urban inputs. This framing positions LLM-based neighborhood-level mobility prediction as a distinct and underexplored problem, asking whether models that have shown promise on individual mobility tasks can also recover collective mobility patterns across neighborhoods from sociodemographic and built-environment context.

In addition, even if LLMs can recover aggregate neighborhood-level outcomes, prediction accuracy alone does not establish that they have learned the empirical relationships that structure mobility across places. Plausible performance may instead reflect shallow heuristics, broad stereotypes \cite{manvi2024large, moayeri2024worldbench}, or memorization \cite{hartmann2023sok} of patterns seen during pretraining, all of which could align with outcomes on average while diverging from the specific relationships that actually structure how context shapes mobility. Thus, whether LLM predictions recover the structural relationships between neighborhood context and travel patterns that empirical data reveal remains an open question.

To address these gaps, we evaluate whether LLMs encode useful priors about urban mobility by testing their ability to infer neighborhood-level mobility outcomes from urban context, assessing the structural alignment of those predictions with empirical context-mobility relationships, and examining whether this alignment holds across cities. Our analysis is grounded in large-scale, real-world mobility trajectories from the United States, provided by Cuebiq, which amasses anonymized location data from nearly 70 million mobile devices, covering roughly 20\% of the U.S. population. We aggregate these data to the Census Block Group (CBG) level and derive multiple mobility indicators capturing three complementary dimensions of behavior: (i) point-level, (ii) trajectory-level, and (iii) temporal-level mobility. Given built-environment and neighborhood sociodemographic characteristics as input, we ask two research questions:

\textbf{RQ1:} To what extent can LLMs predict neighborhood-level mobility outcomes given built-environment and sociodemographic context?

\textbf{RQ2:} To what extent do LLM predictions align with empirically observed relationships between neighborhood characteristics and mobility behavior?

The remainder of the paper is organized as follows. Section~\ref{sec:related_work} reviews prior work on LLMs for mobility generation, prediction, and urban reasoning, as well as recent efforts to evaluate geospatial knowledge and bias in LLMs. Section~\ref{sec:method} presents our problem formulation, mobility outcome construction, zero-shot LLM prediction framework, supervised baselines, and directional alignment methodology. Section~\ref{sec:experiments} describes the datasets, study areas, contextual predictors, model settings, as well as the predictive performance results for RQ1 and the directional alignment results for RQ2. Finally, Section~\ref{sec:conclusion} summarizes the main findings and discusses their implications for using LLMs in aggregate urban mobility inference.

\section{Related Work}
\label{sec:related_work}

\subsection{LLMs for Mobility Generation, Prediction, and Urban Tasks}
A growing body of work uses large language models (LLMs) to generate, predict, and simulate mobility and urban phenomena. In mobility modeling, several studies adapt LLMs to represent individual trajectories, activity chains, and travel behavior. ~\citet{ma2025learning} propose a foundation model for universal human mobility patterns that fuses cross-domain data and semantically enriches GPS traces with survey-informed knowledge, achieving robust performance in activity inference and POI classification across settings such as Los Angeles and Egypt. Wang et al.~\cite{wang2024large} introduce \textsc{LLMob}, an LLM-agent framework that generates personal mobility trajectories by extracting patterns from historical traces and reasoning about daily motivations, producing realistic movement sequences even during abnormal periods such as the pandemic. ~\citet{shao2024beyond} similarly explore the use of LLMs for human mobility generation through context-aware reasoning, emphasizing that language models can explain the intentions behind movement segments rather than merely imitate observed sequences. ~\cite{li2024geo} propose \textsc{Geo-Llama}, a fine-tuned framework for generating mobility trajectories under explicit spatiotemporal constraints, showing that LLMs can produce coherent synthetic movements while satisfying visit-level requirements.

Other work focuses on generating richer behavioral records, such as travel diaries and activity schedules. ~\citet{li2024more} develop \textsc{MobAgent}, which constructs fine-grained profiles and uses recursive reasoning to generate realistic travel diaries aligned with empirical mobility patterns and road-network constraints. \citet{bhandari2024urban} evaluate whether LLMs can generate synthetic travel survey data resembling the National Household Travel Survey (NHTS), finding that fine-tuned models effectively capture complex activity chains and transition probabilities across major U.S. metropolitan areas. Liu et al.~\citet{liu2025human} present a retrieval-augmented framework with a feedback loop for generating daily activity chains under limited information, showing that the approach can reproduce coordinated household behaviors such as joint shopping trips or shared meal times. Extending this direction, ~\citet{liu2025aligning} studies discrete travel demand modeling with persona-conditioned LLMs, demonstrating that behavioral traits and socioeconomic context improve simulation of individual travel choices.

A related stream uses LLMs for mobility prediction and recommendation rather than full-sequence generation. \citet{li2024large} propose \textsc{LLM4POI}, a next-POI recommendation framework that fine-tunes LLMs on check-in data and combines this with key-query similarity to retain collaborative and contextual information. ~\citet{gong2024mobility} introduce \textsc{Mobility-LLM}, a reprogramming framework that uses behavioral prompts and semantic location representations to improve next-location prediction, arrival-time estimation, and user-link prediction. ~\citet{chen2025enhancing} develop \textsc{QT-Mob}, which introduces semantic location tokenization to encode coordinates as semantically meaningful discrete tokens, substantially improving next-location prediction and trajectory recovery. At the urban scale, ~\citet{li2024urbangpt} present \textsc{UrbanGPT}, which integrates spatiotemporal encoding with instruction tuning to align urban interdependencies with LLM reasoning, demonstrating strong zero-shot generalization across prediction tasks such as taxi flows, bike flows, and crime rates.

Taken together, this literature shows that LLMs can serve as flexible models for synthetic mobility generation, travel simulation, recommendation, and urban prediction. It also suggests that these models encode nontrivial mobility-relevant signals. There are a few aggregate-oriented examples: \textsc{UrbanGPT} studies regional spatiotemporal prediction tasks such as taxi flows and bike flows, and \citet{bhandari2024urban} evaluate generated surveys using aggregate pattern-level mobility metrics. Still, the dominant targets of evaluation remain individual trajectories, travel diaries, next-location prediction, or task-specific urban outputs rather than direct recovery of neighborhood-level mobility indicators from contextual urban features, which we address in RQ1. 
\subsection{Evaluating Geospatial and Mobility Knowledge in LLMs}

A second line of work asks what kinds of geographic and mobility knowledge LLMs already encode. ~\citet{manvi2023geollm} introduce \textsc{GeoLLM}, which fine-tunes LLMs on reverse-geocoded map data and shows that the resulting models can accurately predict geospatial socioeconomic indicators such as population density and asset wealth. ~\citet{luo2024deciphering} propose \textsc{TSI-LLM}, a framework for trajectory semantic inference that uses context-rich prompting and chain-of-thought reasoning to infer occupation categories, activity sequences, and detailed textual descriptions from raw movement traces. ~\citet{asano2025mobqa} introduce \textsc{MobQA}, a benchmark of 5,800 question-answer pairs for evaluating LLMs on mobility-related factual retrieval, semantic inference, and interpretive explanation. Their results suggest that while models perform well on straightforward factual extraction, they struggle with more complex semantic reasoning over long GPS trajectories.

This evaluation-oriented literature also highlights important limitations and biases in LLMs' geographic knowledge. ~\citet{manvi2024large} show that although LLMs can make reasonably accurate zero-shot geospatial predictions for objective topics, they exhibit systematic geographic bias on subjective judgments, rating wealthier regions more favorably on attributes such as intelligence or work ethic. ~\citet{wu2024popular} probe demographic and geographic biases in LLM-predicted POI visits, finding that models reflect and amplify race and gender stereotypes in the types of places they associate with different groups. ~\citet{moayeri2024worldbench} further demonstrates broad disparities in factual recall through \textsc{WorldBench}, showing that state-of-the-art LLMs perform substantially worse for countries in low-income and non-Western regions than for Western countries.

Recent work has also begun evaluating LLMs as broader models of urban knowledge rather than only as task-specific predictors. ~\citet{zhang2025genai} introduce \textsc{AI4US}, which tests whether LLMs can reproduce foundational urban-science relationships such as scaling laws, distance decay, and urban vitality. They find that LLMs often recover broad aggregate patterns with high fidelity, but also oversimplify urban complexity, showing limited diversity and weaker causal depth than real urban systems.

Our RQ2 work is most closely related to this evaluation-oriented literature and extends it in a new direction. Rather than examining factual geospatial recall, mobility question answering, or broad urban-theory reproduction, we probe whether LLMs encode empirically meaningful priors about neighborhood-level mobility behavior from built-environment and sociodemographic context.

\begin{figure*}[!htbp]
    \centering
    \includegraphics[width=1\linewidth]{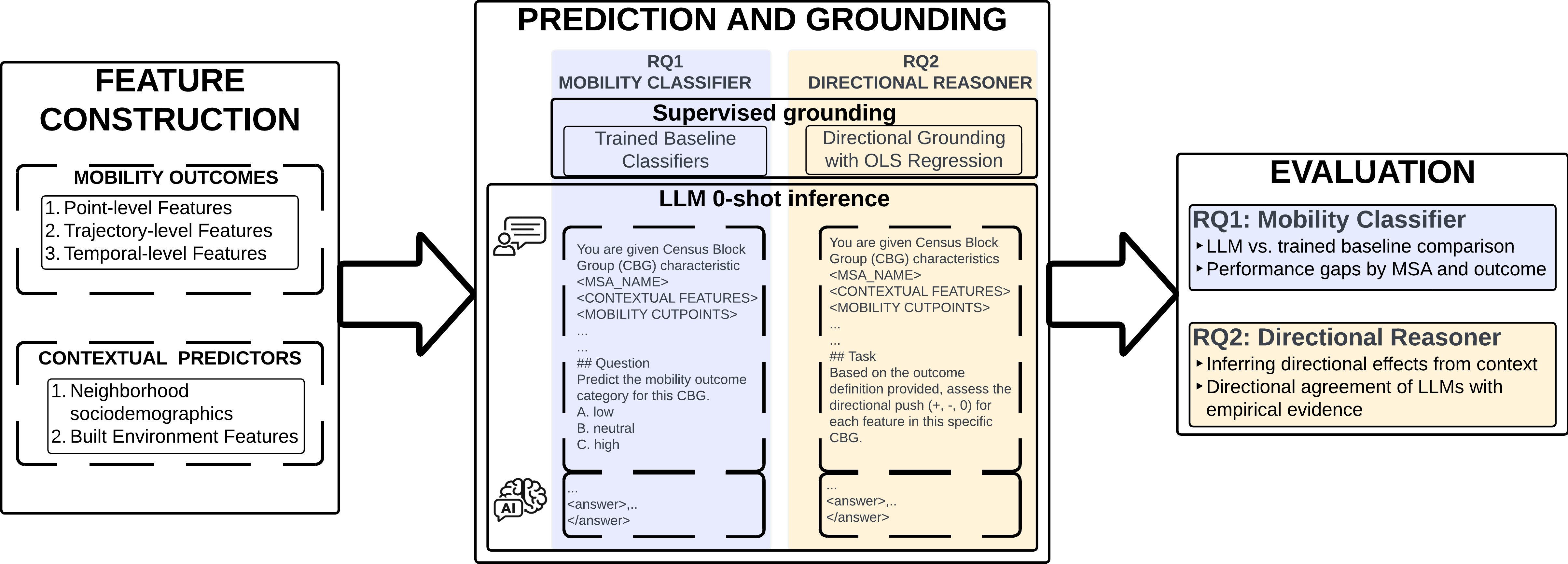}
    \vspace{-1em}
    \caption{Overview of the evaluation pipeline for assessing LLM priors about neighborhood-level mobility.}
    \label{fig:overall_pipeline}
    
\end{figure*}

\section{Method}
\label{sec:method}

\subsection{Problem Formulation}

Let \(i\) index Census Block Groups (CBGs), let \(\mathbf{x}_i \in \mathbb{R}^p\) denote the vector of contextual predictors for CBG \(i\), and let \(m \in \mathcal{M}\) denote a mobility outcome. For each outcome \(m\), let \(z_i^{(m)} \in \mathbb{R}\) denote the continuous CBG-level mobility value derived from aggregated individual mobility traces. We formulate CBG-level mobility outcome prediction as an outcome-specific three-class classification task in which the goal is to infer a discretized mobility label
\begin{equation}
y_i^{(m)} \in \{\texttt{low}, \texttt{neutral}, \texttt{high}\}
\end{equation}
from neighborhood context \(\mathbf{x}_i\). More formally, for each outcome \(m\), we evaluate a predictor
\begin{equation}
F^{(m)} : \mathbf{x}_i \mapsto y_i^{(m)}.
\end{equation}

Here, \(F^{(m)}\) may represent either an LLM-based zero-shot predictor, which reasons over textual descriptions of neighborhood context and outcome semantics, or a supervised baseline trained directly on labeled CBG examples. This formulation allows us to test whether socio-demographic and built-environment context alone is sufficient to recover structured variation in aggregate mobility behavior across neighborhoods.

Our analysis considers two complementary evaluation objectives. The first is \textit{predictive performance (RQ1)}: whether a model can correctly classify the mobility level of a CBG for a given outcome \(m\). The second is \textit{directional alignment (RQ2)}: whether the model captures empirically meaningful relationships between contextual predictors and mobility outcomes. Specifically, for contextual predictor \(x_{ij}\) and mobility outcome \(m\), we assess whether the directional implication of the model’s reasoning agrees with the empirical direction estimated from observed data. This second objective is important because accurate class prediction alone does not establish that a model has recovered the underlying monotonic structure relating urban context to mobility behavior.

The remainder of this section presents the methodological components required to instantiate this formulation. Figure~\ref{fig:overall_pipeline} provides an overview of the evaluation pipeline for assessing LLM priors about neighborhood-level mobility. We first describe how CBG-level mobility outcomes are constructed from individual mobility traces, then introduce the prediction framework for both LLM-based and supervised models, and finally present the directional alignment framework used to compare model-implied relationships with empirical patterns in the data.

\subsection{Mobility Outcome Construction}

In this paper, we construct CBG-level mobility outcomes from anonymized individual mobility traces. The goal is to derive aggregate measures that characterize different dimensions of resident mobility behavior and that can be paired with socio-demographic and built-environment predictors at the same spatial scale. 

\subsubsection{From Individual Mobility Data to CBG-Level Aggregates}

The mobility data used in this study consist of anonymized individual mobility traces collected over 2021. From these traces, we first derive user-level mobility measures described in \ref{sec:def}, that summarize annual movement behavior. These measures are computed from observed stops and travel episodes and capture complementary aspects of mobility, including spatial extent, movement structure, and temporal variability.

Because our prediction task is defined at the CBG level, we aggregate the user-level mobility measures to the CBG scale. For each user, we identify the associated home CBG and assign the user-level mobility measures to that area. Then, for each CBG, we summarize the mobility behavior of all associated users to obtain aggregate mobility outcomes representing the typical mobility profile of residents in that CBG.

This aggregation serves two purposes. First, it aligns the mobility outcomes with the same spatial unit used for the contextual predictors in the classification task. Second, it preserves privacy by ensuring that the prediction targets are aggregate behavioral summaries rather than individual trajectories. To improve reliability, we retain only CBGs with sufficient user support in the mobility data. The resulting dataset therefore, consists of robust CBG-level mobility outcomes that serve as the ground-truth targets throughout the paper.

\subsubsection{Mobility Outcome Families and Definitions}
\label{sec:def}

\begin{table*}[t]
\centering
\footnotesize
\label{tab:mobility_outcomes}
\begin{tabular}{llll}
\toprule
\textbf{Family} & \textbf{Outcome} & \textbf{What it measures} & \textbf{Interpretation of higher values} \\
\midrule

\multirow{4}{*}{Point-level}
& Stay-points entropy
& Evenness of dwell time across visited locations
& Activity time is distributed more evenly across places \\

& Radius of gyration
& Dispersion of stops around the center of activity
& Larger spatial spread of routine activity locations \\

& Convex hull diameter
& Maximum distance across the activity space
& Wider geographic extent of observed mobility \\

& Ellipse area
& Size of elliptical approximation of activity space
& Broader spatial footprint of activity locations \\

\midrule

\multirow{3}{*}{Trajectory-level}
& Total travel length
& Cumulative distance traveled across observed trips
& More extensive travel over the observation period \\

& Travel entropy
& Diversity of route or trip signatures
& More varied routing or travel behavior \\

& Average duration
& Mean duration of observed travel segments
& Longer average trips \\

\midrule

Temporal-level
& Daily temporal fragmentation
& Within-day variability in dwell durations
& Greater irregularity in daily activity timing \\

\bottomrule
\end{tabular}

\caption{Summary of mobility outcome families and metrics. All metrics are computed at the user-year level from annual stop and travel-segment sequences and then aggregated to the Census Block Group (CBG) level.}
\end{table*}

Following \citet{wu2019inferring} taxonomy of mobility characterization from individual trajectories, we organize the mobility outcomes into three families: \textbf{point-level}, \textbf{trajectory-level}, and \textbf{temporal-level} outcomes. Point-level outcomes summarize the spatial footprint and organization of visited locations; trajectory-level outcomes characterize movement over observed travel sequences; and temporal-level outcomes quantify the temporal regularity of activity patterns.

For each user, we represent annual mobility as an ordered sequence of stops and travel segments,
\begin{equation}
s_1 \xrightarrow{\mathrm{trv}_1} s_2 \xrightarrow{\mathrm{trv}_2} \cdots \xrightarrow{\mathrm{trv}_{N-1}} s_N,
\end{equation}
where each stop \(s_i\) is associated with a location \(\mathbf{x}_i \in \mathbb{R}^2\), a dwell time \(\tau_i\), and an observation day \(d_i\), and each travel segment \(\mathrm{trv}_i\) connecting \(s_i\) and \(s_{i+1}\) is associated with a travel length \(L_i\) and a travel duration \(a_i\). Unless otherwise noted, all outcomes are first computed at the user-year level and then aggregated to the Census Block Group (CBG) level. We further define normalized dwell-time weights as
\begin{equation}
w_i = \frac{\tau_i}{\sum_{k=1}^{N} \tau_k}.
\end{equation}

\paragraph{(1) Point-level outcomes.}
Point-level outcomes quantify the extent, dispersion, and spatial organization of visited locations. Let a user have stops indexed by \(i=1,\dots,n\), where stop \(i\) has coordinates \(\mathbf{s}_i\) and dwell time \(d_i\). Define normalized dwell-time weights as
\begin{equation}
w_i = \frac{d_i}{\sum_{k=1}^{n} d_k}.
\end{equation}

\textit{Stay-points entropy} measures how evenly dwell time is distributed across distinct visited locations. Let \(\mathcal{L}\) denote the set of unique stay locations, and let \(p_l\) be the proportion of total dwell time spent at location \(l \in \mathcal{L}\). Then
\begin{equation}
H_{\text{stay}} = - \sum_{l \in \mathcal{L}} p_l \log_2 p_l.
\end{equation}
Higher values indicate that activity time is distributed more evenly across locations, whereas lower values indicate concentration in a small number of places. 

\textit{Radius of gyration} measures the characteristic distance of visited locations from the user’s center of activity. Let
\begin{equation}
\bar{\mathbf{x}} = \sum_{i=1}^{N} w_i \mathbf{x}_i
\end{equation}
denote the weighted centroid, and let \(\delta(\mathbf{x}_i,\bar{\mathbf{x}})\) denote the haversine distance between stop \(i\) and the centroid. The radius of gyration is
\begin{equation}
r_g = \sqrt{\sum_{i=1}^{N} w_i \, \delta(\mathbf{x}_i,\bar{\mathbf{x}})^2 }.
\end{equation}
Larger values indicate more spatially dispersed activity patterns.

\textit{Convex hull diameter} quantifies the maximum spatial extent of the activity space. Let \(\mathcal{H}\) denote the set of vertices of the convex hull formed by the observed stop locations. Then,
\begin{equation}
D_{\text{hull}} = \max_{\mathbf{u},\mathbf{v} \in \mathcal{H}} d(\mathbf{u},\mathbf{v}),
\end{equation}
where \(d(\cdot,\cdot)\) is the haversine distance. This outcome represents the largest distance between any two locations in a user’s activity space.

\textit{Ellipse area} measures the size of an elliptical approximation of the activity space. Let
\begin{equation}
\mathbf{C} = \sum_{i=1}^{n} w_i (\mathbf{s}_i-\bar{\mathbf{s}})(\mathbf{s}_i-\bar{\mathbf{s}})^\top
\end{equation}
be the weighted covariance matrix of stop coordinates, and let \(\lambda_1 \geq \lambda_2\) be its eigenvalues. The ellipse area is
\begin{equation}
A_{\text{ellipse}} = \pi \sqrt{\lambda_1}\sqrt{\lambda_2}\, c^2,
\end{equation}
where \(c\) is a scale factor. Larger values indicate a broader activity space.

\paragraph{(2) Trajectory-level outcomes.}
Trajectory-level outcomes characterize mobility over sequences of trips and capture the extent and diversity of travel behavior.

\textit{Total travel length} measures the cumulative distance traveled over all observed trips. Let trajectory segments be indexed by \(t=1,\dots,T\), and let \(L_t\) denote the length of segment \(t\). Then,
\begin{equation}
L_{\text{total}} = \sum_{t=1}^{T} L_t.
\end{equation}
Higher values indicate more extensive travel over the observation period.

\textit{Travel entropy} measures the diversity of routes used by a user. Let \(\mathcal{R}\) denote the set of unique route signatures, and let \(p_r\) be the proportion of trips corresponding to route \(r \in \mathcal{R}\). Then,
\begin{equation}
H_{\text{travel}} = - \sum_{r \in \mathcal{R}} p_r \log_2 p_r.
\end{equation}
Higher values indicate more varied routing behavior, while lower values indicate repetitive travel patterns.

\textit{Average duration} measures the mean duration of observed trips. Let \(a_i\) denote the duration of travel segment \(\mathrm{trv}_i\), \(i=1,\dots,N-1\). Then
\begin{equation}
\bar{a} = \frac{1}{N-1}\sum_{i=1}^{N-1} a_i.
\end{equation}
Higher values indicate longer average trip durations over the observation period.

\paragraph{(3) Temporal-level outcomes.}
Temporal-level outcomes characterize the temporal organization and irregularity of mobility behavior.

\textit{Daily temporal fragmentation} measures within-day variability in dwell durations. For each observed day \(d \in \mathcal{D}\), let
\begin{equation}
\sigma_d^2 = \mathrm{Var}\{\tau_i : d_i = d\}
\end{equation}
denote the variance of dwell times across all stops observed on that day. We define daily temporal fragmentation as
\begin{equation}
F_{\text{daily}} = \frac{1}{|\mathcal{D}|}\sum_{d \in \mathcal{D}} \sigma_d^2.
\end{equation}
Higher values indicate greater within-day temporal irregularity, reflected in a wider mix of short and long stays across stops within observed days.

These three types of mobility outcomes provide complementary views of aggregate mobility behavior at the CBG level. In the remainder of the paper, we use this organization to structure the prediction tasks, the directional alignment analysis, and the presentation of results.

\subsection{Mobility Outcome Prediction Framework}
\label{sec:prediction_framework}

We evaluate two classes of predictors for CBG-level mobility outcome prediction: \textbf{LLM-based zero-shot predictors} and \textbf{supervised baseline models}. Both use the same contextual predictors for each CBG, but they differ in how the mapping from neighborhood context to mobility outcomes is obtained. The LLM setting evaluates zero-shot contextual inference from textual descriptions, whereas the supervised baselines learn the mapping directly from labeled examples.

Recall from Section~\ref{sec:def} that each CBG \(i\) is represented by a contextual predictor vector \(\mathbf{x}_i \in \mathbb{R}^p\), and that for each mobility outcome \(m \in \mathcal{M}\), the prediction target is the discretized class label \(y_i^{(m)} \in \{\texttt{low}, \texttt{neutral}, \texttt{high}\}\). The prediction task is therefore outcome-specific: for each \(m\), we construct a predictor \(F^{(m)}\) that maps neighborhood context \(\mathbf{x}_i\) to a three-class mobility label.

\subsubsection{LLM-Based Prediction}

For each CBG \(i\) and mobility outcome \(m\), we construct an outcome-specific prompt \(P(i,m)\) that describes the local contextual characteristics of the CBG and asks the model to predict the corresponding mobility class (0-shot). Each prompt contains four components (see Figure~\ref{fig:rq1_user_prompt}):  
(1) the Core-Based Statistical Area (CBSA) in which the CBG is located;  
(2) a natural-language definition of the target mobility outcome \(m\);  
(3) the CBSA-specific tertile thresholds used to discretize the continuous outcome into \texttt{low}, \texttt{neutral}, and \texttt{high} classes; and  
(4) a textual profile of the CBG’s contextual predictors.
The associated system prompt is provided in the Appendix, Figure~\ref{fig:rq1_sys_prompt}.

\begin{figure}[H]
\begin{tcolorbox}[
  colback=gray!8,
  colframe=black!50,
  boxrule=0.5pt,
  title={\small\bfseries User Prompt Template},
  left=1pt,
  right=1pt
]
\scriptsize
\begin{verbatim}
You are given Census Block Group (CBG) characteristics for CBSA: [CBSA name].

## Outcome Description
Mobility outcome: [outcome name]
Definition: [outcome definition]

Tertile cutpoints within [CBSA name]: low <= [c1]; neutral ([c1], [c2]];
high > [c2].

## Predictor Descriptor
- Percentiles are computed within the CBSA.
- Predictor buckets correspond to CBSA-wide quintiles:
  very_low | low | neutral | high | very_high.
- Outcome labels are tertiles: low | neutral | high.

## CBG Features
- [feature 1]; [raw value] (CBSA percentile Pxx, [bucket])
- [feature 2]; [raw value] (CBSA percentile Pxx, [bucket])
- ...
- [feature p]; [raw value] (CBSA percentile Pxx, [bucket])

## Question
Predict the mobility outcome category for this CBG.
A. low
B. neutral
C. high
\end{verbatim}
\end{tcolorbox}
\vspace{-1.5em}
\caption{User prompt for RQ1.}
\label{fig:rq1_user_prompt}
\vspace{-1.5em}
\end{figure}

The contextual predictors are expressed in a percentile-based format designed to convey both absolute and relative neighborhood context. For each predictor \(x_{ij}\), the prompt includes its raw value, its within-CBSA percentile rank \(r_{ij}\), and a quintile-based categorical bucket
\[
b_{ij} \in \{\texttt{very\_low}, \texttt{low}, \texttt{neutral}, \texttt{high}, \texttt{very\_high}\}.
\]
Thus, each CBG is represented as a structured textual profile summarizing how its socio-demographic and built-environment characteristics compare with those of other CBGs in the same metropolitan area. This prompt design encourages the model to reason over relative urban context rather than relying only on raw feature magnitudes.

The task is outcome-specific. For each mobility outcome \(m\), a separate prompt is constructed using the outcome definition and the corresponding CBSA-specific tertile thresholds, while the contextual predictor profile of the CBG remains fixed. The model is then asked to output one of three class labels, $\{\texttt{low}, \texttt{neutral}, \texttt{high}\}$. In addition to the class prediction, the model is instructed to provide brief reasoning and a ranked assessment of influential predictors. 

For comparison, we also train a set of supervised baseline models on the same contextual predictors and outcome labels. Specifically, for each mobility outcome \(m\), we fit the following multiclass classifiers: \texttt{logit\_multinomial}, \texttt{random\_forest}, \texttt{decision\_tree}, \texttt{hist\_gradient\_boosting}, and \texttt{xgboost}. These trained baselines provide data-driven reference points for the classification task and allow us to compare zero-shot contextual inference by LLMs against conventional supervised learning approaches trained directly on labeled CBG examples.





\subsubsection{Evaluation Setup}

Prompts are generated for all available CBGs and mobility outcomes. For predictive evaluation, however, we report LLM performance on the same held-out \(20\%\) test split used for the supervised baselines, so that both model classes are evaluated on an identical subset of CBGs allowing a direct comparison between zero-shot LLM-based contextual inference and conventional supervised prediction for CBG-level mobility outcome classification.

\subsection{Directional Alignment Framework}
\label{sec:directional_alignment}

Since classification accuracy alone does not indicate whether a model captures the empirical structure relating neighborhood context to mobility behavior, we also examine whether correct predictions may still be based on misleading or weakly grounded reasoning. To address this, we introduce a complementary \emph{directional alignment} analysis that evaluates whether the directional implications of model reasoning (LLM) are consistent with empirical relationships observed in the data (baseline models).

Let \(x_{ij}\) denote contextual predictor \(j\) for CBG \(i\), and let \(z_i^{(m)}\) denote the continuous value of mobility outcome \(m\). For each CBSA \(c\), mobility outcome \(m\), and contextual predictor \(j\), 
we estimate a ground truth empirical directional relationship between \(x_{ij}\) and \(z_i^{(m)}\) as follows. 
We fit an ordinary least squares (OLS) regression in which the continuous mobility outcome is regressed on the contextual predictors. The sign of the standardized coefficient for predictor \(j\), denoted \(\beta_{cjm}^{\text{OLS}}\), provides a parametric estimate of direction:
\begin{equation}
\mathrm{dir}_{cjm}^{\text{OLS}} = \mathrm{sign}(\beta_{cjm}^{\text{OLS}}) \in \{-1,+1\}.
\end{equation}


We then derive an LLM-implied directional signal from the model outputs. As shown in the system prompt Figure~\ref{fig:rq2_sys_prompt} and corresponding user prompt  Figure~\ref{fig:rq2_user_prompt} in Appendix, the model is additionally asked to provide a ranked assessment of influential predictors together with whether each predictor pushes the mobility outcome downward, upward, or neither. We encode this directional push as $p_{ijm} \in \{-1,0,+1\},$ where \(-1\), \(0\), and \(+1\) denote negative, neutral, and positive influence on outcome \(m\), respectively.



To test whether the LLM-implied pushes agree with the empirical OLS direction, we examine how \(p_{ijm}\) varies across the ordered predictor buckets (described using a quintile-based bucket, $
b_{ij} \in \{\texttt{very\_low}, \texttt{low}, \texttt{neutral}, \texttt{high}, \texttt{very\_high}\}
$). If the empirical coefficient for predictor \(j\) is positive, then increasing values of \(b_{ij}\) should correspond to increasingly positive LLM pushes. If the empirical coefficient is negative, then increasing values of \(b_{ij}\) should correspond to increasingly negative LLM pushes. We formalize this by orienting each LLM-implied push by the empirical OLS sign:
\begin{equation}
\tilde{p}_{ijm} = \mathrm{dir}_{cjm}^{\text{OLS}} \times p_{ijm}.
\end{equation}
After this transformation, \textbf{larger values of \(\tilde{p}_{ijm}\) always indicate stronger agreement with the empirical direction}, regardless of whether the OLS coefficient is positive or negative.

For each CBSA--predictor--outcome combination \((c,j,m)\), we then apply two one-sided Jonckheere--Terpstra (JT) tests to \(\tilde{p}_{ijm}\) across the ordered buckets \(b_{ij}\). The \textit{increasing JT test} evaluates whether the OLS-oriented LLM signal becomes more aligned with the empirical direction as the predictor bucket increases from \texttt{very\_low} to \texttt{very\_high}. The \textit{decreasing JT test} evaluates whether the LLM signal instead moves in the opposite direction. We apply Benjamini--Hochberg correction within each CBSA--outcome panel to account for multiple predictors.

Based on the corrected JT results, we assign each \((c,j,m)\) combination one of three directional labels:
\[
\{\texttt{aligned\_increasing},\ 
\texttt{opposite\_trend},\ 
\texttt{no\_clear\_trend}\}.
\]
A relationship is labeled \texttt{aligned\_increasing} when the increasing JT test is significant and the decreasing test is not, indicating that the LLM-implied direction agrees with the empirical OLS direction. It is labeled \texttt{opposite\_trend} when the decreasing JT test is significant and the increasing test is not, indicating that the LLM-implied direction moves against the empirical relationship. All remaining cases are labeled \texttt{no\_clear\_trend}. 

\section{Experiments}
\label{sec:experiments}

\subsection{Experimental Setting}
\label{sec:exp_setting}

\begin{table}[t]
\centering
\footnotesize
\label{tab:cbsa_device_summary}
\begin{tabular}{lrrrr}
\toprule
CBSA & \# CBGs & Total devices & Mean devices/CBG & Median [Q1, Q3] \\
\midrule
ATL        & 1051 & 209099 & 198.95 & 146 [89, 266] \\
LA      & 4196 & 359987 & 85.79 & 73 [52, 104] \\
MIA & 1730 & 273187 & 157.91 & 126 [80, 195] \\
SF       & 1779 & 134379 & 75.54 & 57 [37, 90] \\
\bottomrule
\end{tabular}

\caption{Coverage statistics across the study CBSAs. The table reports the number of CBGs covered, the total number of devices, and the distribution of devices per CBG.}
\label{tab:coverage_stats}
\vspace{-2em}
\end{table}




\subsubsection{Data}
\label{sec:exp_data}
\paragraph{Mobility data.}
We derive the mobility outcomes from anonymized and aggregated mobility traces provided by Cuebiq\footnote{\url{https://docs.spectus.ai/}}. Aggregated mobility data is provided by Cuebiq, a location intelligence platform. Data is collected from anonymized users who have opted-in to provide access to their location data anonymously, through a CCPA and GDPR-compliant framework. Through its Social Impact program, Cuebiq provides mobility insights for academic research and humanitarian initiatives. The Cuebiq responsible data sharing framework enables research partners to query anonymized and privacy-enhanced data, by providing access to an auditable, on-premise Data Cleanroom environment. All final outputs provided to partners are aggregated in order to preserve privacy.
 The data cover four Core-Based Statistical Areas (CBSAs) in the United States: Atlanta--Sandy Springs--Roswell, GA \textbf{(ATL)}; Los Angeles--Long Beach--Anaheim, CA \textbf{(LA)}; Miami--Fort Lauderdale--West Palm Beach, FL \textbf{(MIA)}; and San Francisco--Oakland--Berkeley, CA \textbf{(SF)}. We use data from calendar year 2021 and retain only users observed on more than 60 days during the year. To ensure reliable CBG-level aggregation, we further restrict the analysis to CBGs with at least 20 devices. Under these criteria, the final dataset covers 8,756 CBGs \footnote{CBGs corresponding to military locations are excluded in accordance with Cuebiq's data-sharing policies.} and 976,652 devices across the four CBSAs. At the CBG level, device support varies across study areas, with median devices per CBG ranging from 57 in SF to 144 in ATL. These mobility traces are used to construct the continuous user-level mobility measures described in Section~\ref{sec:def}, which are then aggregated to the CBG level and discretized into outcome classes for prediction. Table~\ref{tab:coverage_stats} and Figure~\ref{fig:quantile_plots} (Appendix) show the distribution of the sample sizes as well as the CBGs covered. 

\paragraph{Socio-demographic predictors.}
We obtain socio-demographic variables from the 2019 American Community Survey (ACS) 5-year estimates. These predictors characterize the population and household composition of each CBG and include measures such as median income, racial and ethnic composition, age structure, educational attainment, and vehicle ownership. In the prediction framework, these variables serve as contextual signals describing the demographic and socioeconomic profile of each neighborhood. For the OLS, we drop some of the correlated features. 

\paragraph{Built-environment predictors.}
Built-environment variables are derived from the EPA Smart Location Database (SLD), which provides CBG-level indicators related to density, street-network structure, transit accessibility, and land-use mix. Because the raw SLD variables are numerous and highly correlated, we summarize them using principal component analysis (PCA). For each conceptual built-environment domain, we retain the first principal component and use it as a composite index in the downstream analysis. This yields four interpretable built-environment predictors corresponding to density, connectivity, transit access, and land-use mix.

\paragraph{Predictor standardization.}
To support both the LLM prompting framework and the supervised baselines, all predictors are harmonized at the CBG level. For the LLM-based experiments, each predictor is represented using its raw value, within-CBSA percentile, and a quintile-based bucket. For the supervised baselines, the same underlying predictor values are used directly as numeric features.

\subsubsection{LLM Prediction Settings}
\label{sec:exp_llm_settings}

We evaluate models from eight families in a zero-shot setting using the prompt
template described in Section~\ref{sec:prediction_framework}: Claude, GPT,
GPT-OSS, Gemini, Gemma, Qwen, DeepSeek, and Llama (Table~\ref{tab:parse_success},
Appendix). The set spans both proprietary and open-weight models across a range
of sizes. Open-weight models are served with \texttt{vLLM}, while proprietary and other hosted models are accessed through AWS cloud infrastructure. Across all models, decoding is deterministic with temperature \(0.0\). For each CBG and
mobility outcome, we issue one outcome-specific prompt and parse the response
into a class prediction and predictor-level directional assessments LLM performance is reported on the same held-out \(20\%\) test split used for the supervised baselines. For the RQ1 accuracy comparison, LLM
performance is reported on the same held-out \(20\%\) test split used for the
supervised baselines. However, for the RQ2 directional analysis, we retain all the CBGs to maximize the power of the JT trend tests.

\subsection{RQ1: Predictive Performance Results}
\label{sec:rq1_results}

\begin{figure*}
    \centering
    \includegraphics[width=0.95\linewidth]{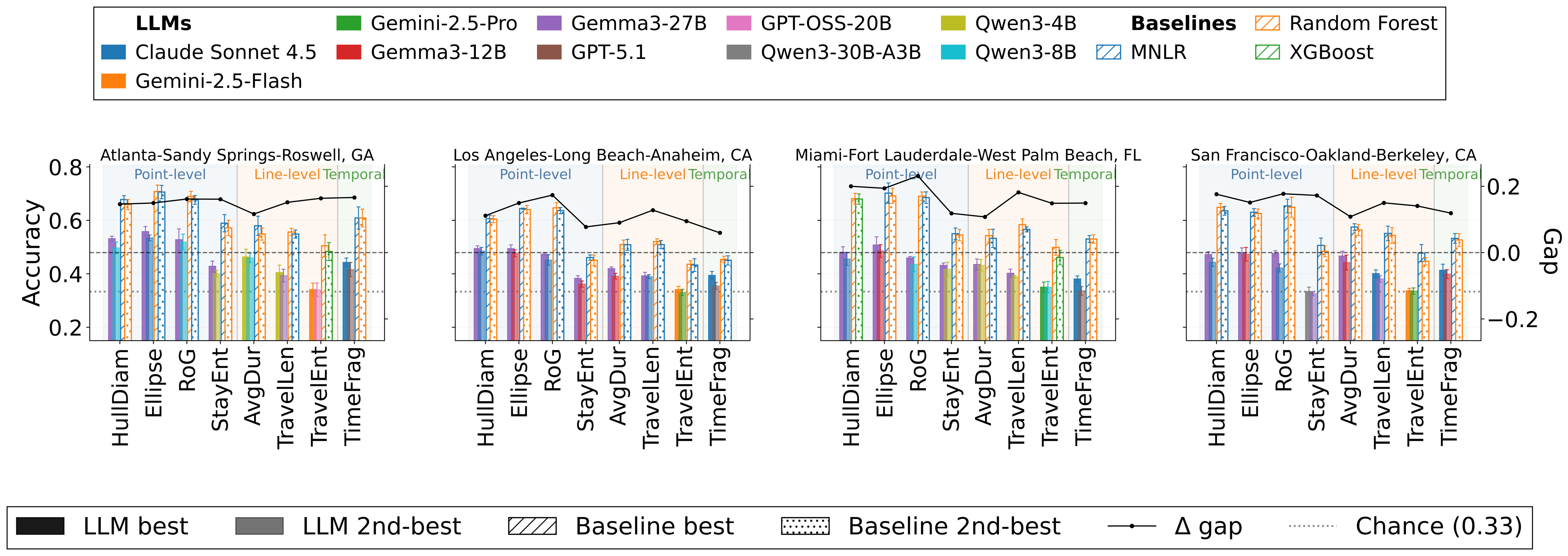}
    \vspace{-1em}
    \caption{Mean 3-class prediction accuracy across four CBSAs for mobility outcomes inferred from socio-demographic and built-environment predictors. Outcomes are grouped into point-level, line-level, and temporal mobility features; bars show the top two LLMs and top two supervised baselines for each outcome, averaged over repeated 20\% held-out test splits, with error bars denoting standard deviation across 5 repetitions. The horizontal dashed line marks chance accuracy (1/3), and the black line shows the performance gap between the best baseline and best LLM for each outcome. Across CBSAs, supervised baselines consistently outperform zero-shot LLMs, although LLMs remain above chance on many tasks, with the strongest performance typically observed for point-level mobility outcomes.}
    \label{fig:RQ1_Results}
    
\end{figure*}

\paragraph{\textbf{Finding 1.} Neighborhood context provides strong signal for CBG-level mobility prediction, but zero-shot LLMs only partially recover it.}


Figure~\ref{fig:RQ1_Results} shows that CBG-level mobility outcomes are learnable from socio-demographic and built-environment context. Across the 32 city--outcome pairs, the best supervised baseline achieves an average accuracy of \(0.580\), well above the random three-class baseline of \(0.333\). This confirms the notion that contextual predictors contain meaningful information about aggregate mobility behavior. Zero-shot LLMs also recover part of this signal, with the best LLM averaging \(0.435\) accuracy across all outcomes. Although LLMs remain consistently below supervised baselines, with an average best-baseline--best-LLM gap of \(0.144\) (range \([0.059\)--\(0.231]\)) accuracy points, this gap is also expected because the supervised models are trained directly on the target mobility labels while the LLMs operate without task-specific examples. Table~\ref{tab:city_outcome_top_models} contains the top performing model results and the their corresponding gaps with the best trained baselines.

\paragraph{\textbf{Finding 2.} Spatial extent outcomes are most predictable, but also exhibit the largest LLM--baseline gaps.}

Figure~\ref{fig:RQ1_Results} shows that the spatial extent measures carry the
strongest signal from neighborhood context and also show the widest gap between
zero-shot LLMs and supervised baselines. For example, for \textit{convex hull diameter}, \textit{ellipse area}, and \textit{radius of gyration}, the best supervised baselines achieve average accuracies of \(0.653\), \(0.671\), and \(0.670\), respectively, while the best LLMs reach \(0.495\), \(0.510\), and \(0.484\). Thus, LLMs perform relatively well on activity-space extent, but supervised models extract substantially more predictive structure from the same contextual features. The average LLM--baseline gap for these three outcomes is \(0.168\), larger than the gap for line-level outcomes (\(0.132\)) and the temporal outcome (\(0.123\)). The pattern is especially pronounced in MIA CBSA, where the gap reaches \(0.231\) for \textit{radius of gyration}, \(0.200\) for \textit{convex hull diameter}, and \(0.194\) for \textit{ellipse area}. 


\paragraph{\textbf{Finding 3:} Entropy-based mobility measures are the least predictable from static neighborhood context.}
Figure~\ref{fig:RQ1_Results} shows that the two diversity measures,
\textit{stay-point entropy} and \textit{travel entropy}, are the weakest
outcomes for zero-shot LLMs and are among the weakest for supervised baselines.
For LLMs they are the only outcomes whose best accuracy falls below \(0.40\),
averaging \(0.394\) (stay-point entropy) and \(0.342\) (travel entropy) across
the four CBSAs. Travel entropy is the more extreme case: the best LLM stays
near the three-class chance level of \(0.333\) in every CBSA, recovering little
context-based signal, whereas stay-point entropy recovers a weak signal in most
cities but also falls to chance in SF. Supervised baselines perform lowest on
the same two outcomes (\(0.480\) and \(0.527\)), the two lowest of any outcome,
so the limitation is not specific to LLMs but reflects weaker predictability
from the available CBG-level socio-demographic and built-environment predictors.
Unlike spatial extent, these entropy measures capture behavioral variability in
which locations and trips are taken, which likely depends on finer-grained
factors such as individual routines, trip purpose, and day-to-day variability
that static neighborhood attributes do not capture.



\paragraph{\textbf{Finding 4.} Gemma3-27B leads across most mobility outcomes, while Claude-Sonnet is strongest for daily temporal fragmentation.}
Across the LLM results in Fig.~\ref{fig:RQ1_Results}, \texttt{Gemma3-27B} is the
most consistent zero-shot predictor. It is the top LLM in 20 of the 32
city--outcome pairs, ranks first for \textit{convex hull diameter},
\textit{ellipse area}, and \textit{radius of gyration} in all four CBSAs, and is
frequently the second-best LLM elsewhere. The exception is \textit{daily
temporal fragmentation}, where \texttt{Claude-Sonnet-4.5} is the strongest LLM
in all four CBSAs and \texttt{GPT-5.1} is second in three of them, so the
proprietary chat-tuned models are comparatively strongest on the temporal
outcome rather than the spatial ones. The best-performing LLM therefore depends
on the mobility dimension being inferred.

\subsection{RQ2: Directional Alignment Results}
\label{sec:rq2_results}

\begin{figure*}[!ht]
    \centering
    \includegraphics[width=0.9\linewidth]{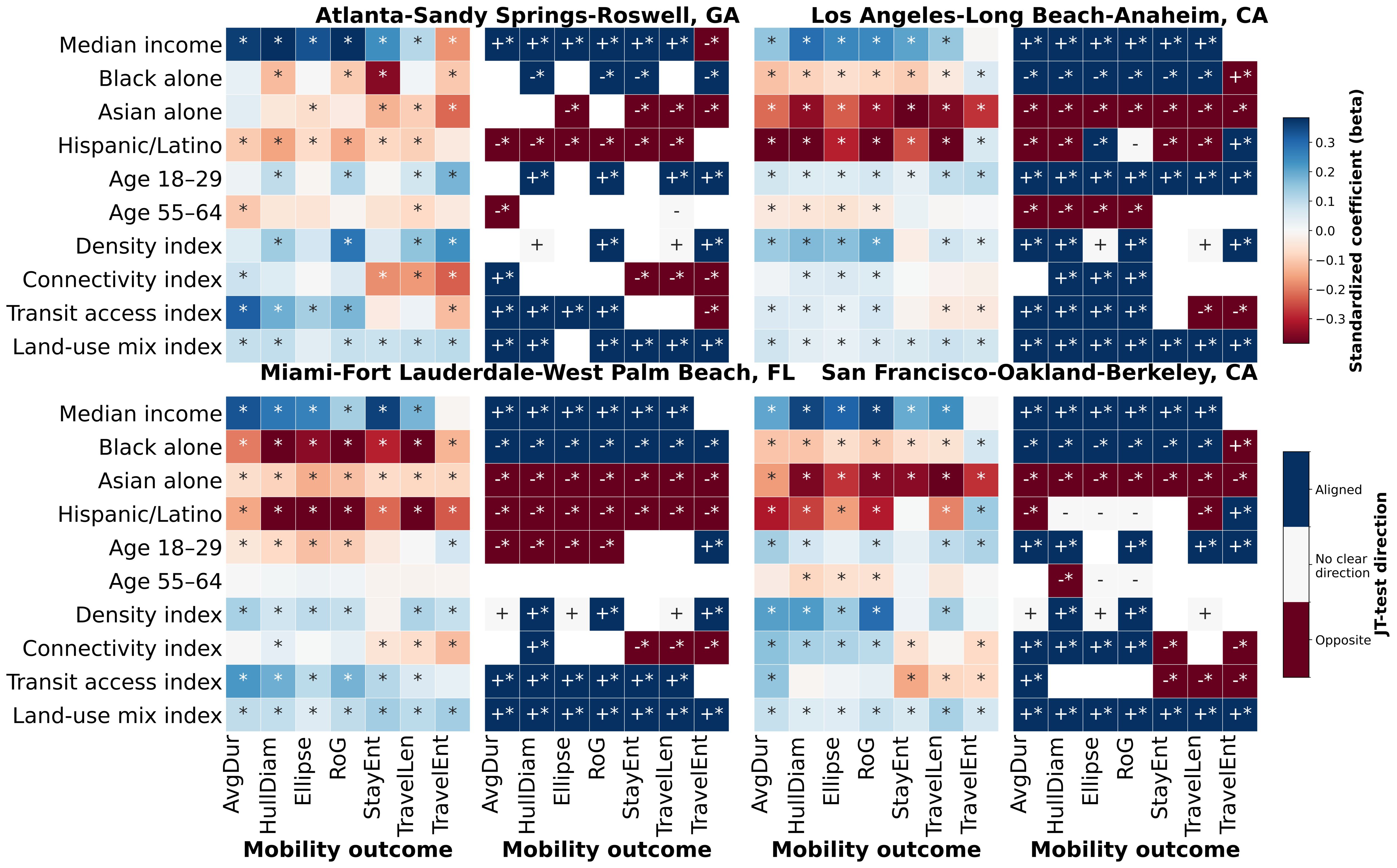}
    \vspace{-1em}
    \caption{Empirical and LLM-implied directional relationships between contextual predictors and mobility outcomes across the four CBSAs. For each CBSA, the left heatmap reports standardized OLS regression coefficients
    The right heatmap reports the corresponding LLM-implied directions from JT trend tests, restricted to predictor--outcome pairs with statistically significant OLS coefficients. 
    Each JT cell is annotated with the corresponding standardized OLS coefficient and JT significance indicator.
    Results are shown for \texttt{Gemma-3-27B}, the best-performing LLM overall across mobility outcomes, except for \textit{Daily Temporal Fragmentation}, for which \texttt{Claude-Sonnet-4.5} results are reported in Appendix Figure~\ref{fig:claude_daily_temporal_fragmentation}.}
    \label{fig:reg_jt_city_alignment}
    \vspace{-1em}
\end{figure*}

\begin{figure}[b]
\includegraphics[width=1\linewidth]{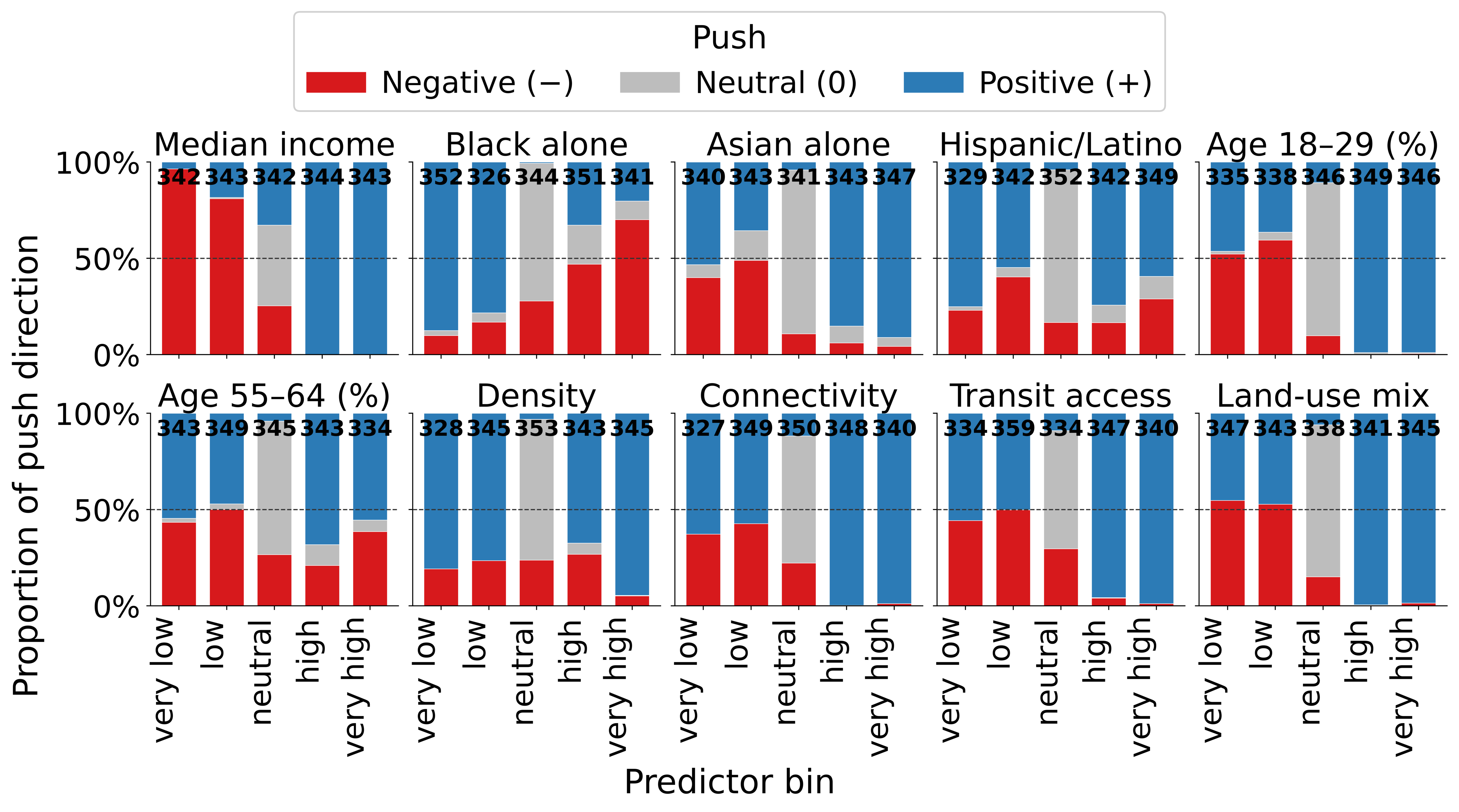}
\vspace{-1.5em}
\caption{Example of LLM prediction shifts across predictor bins for SF CBSA and \textit{Convex Hull Diameter}. Bars show the percentage of LLM predictions pushed toward lower, neutral, or higher mobility classes within each bin, illustrating how the model's directional response is estimated from ordered predictor values. Text annotations report the number of CBGs in each bin.}

 \label{fig:example_llm_push}
 \vspace{-2em}
\end{figure}

RQ2 asks whether LLMs recover empirically meaningful directional relationships between neighborhood context and mobility outcomes. Figure~\ref{fig:reg_jt_city_alignment} compares the empirical directions estimated from OLS regression (left panels) with the LLM-based directional alignment results from the Jonckheere--Terpstra analysis (right panels). Overall, the results suggest that LLMs often capture the \emph{broad sign} of context--mobility relationships, but do so in a coarse and simplified way that does not fully reflect metropolitan variation or outcome-specific complexity. To illustrate how the LLM-implied directions are derived, Figure~\ref{fig:example_llm_push} shows one example for SF CBSA and \textit{Convex Hull Diameter}. 

Because \texttt{Gemma-3-27B} achieves the strongest classification performance for the point-level and line-level mobility outcomes, we report its directional alignment results for those outcome groups. For the only temporal feature, \texttt{Claude-Sonnet-4.5} performs best across CBSAs; we therefore report its directional results separately for that outcome in Figure~\ref{fig:claude_daily_temporal_fragmentation} in the Appendix.

\paragraph{\textbf{Finding 1:} For empirically significant relationships, Gemma-3-27B applies predictor-level directional priors that are largely invariant across mobility outcomes.} 
Among predictor--outcome pairs with statistically significant OLS coefficients, Gemma-3-27B shows a strong tendency to assign a fixed direction to each contextual predictor, rather than adapting its directional expectation to the specific mobility outcome. This pattern is clearest when considering cases where the JT test yields a clear monotonic direction. In these cases, 9/10 contextual predictors receive a single implied direction across nearly all city--outcome combinations. Black population share is the clearest example. Across the 25 Black population share relationships with a clear JT direction, Gemma-3-27B implies a negative direction in all 25 cases. This pattern holds for all seven mobility outcomes in LA, MIA, and SF, and for four of the seven outcomes in ATL, with the remaining ATL outcomes showing no clear monotonic direction.

The same outcome-invariant structure appears for several predictors with positive LLM-implied directions. Median income is mapped to a positive direction in all 25 cases with a clear JT direction. Asian population share is also mapped to a positive direction in all 25 clear cases, despite the empirical OLS coefficients being negative in many of these relationships. Land-use mix shows the most uniform positive pattern, with positive implied directions in 27 of 28 city--outcome pairs and only one no-direction case. Transit access is similarly consistent: all 21 cases with a clear JT direction are positive. Connectivity, density, and the share of young adults also show positive implied directions whenever the model produces a clear monotonic trend, although these predictors have more no-direction cases. Hispanic/Latino population share is the main exception to complete invariance, but it is still predominantly mapped to a positive direction, with 21 positive implied directions, one negative implied direction, and six no-direction cases.

These results suggest that Gemma-3-27B does not primarily reason about how a contextual predictor should affect each mobility feature separately. Instead, once a predictor is salient, the model appears to attach a broad directional prior to that predictor and reuse it across different types of mobility outcomes.

\paragraph{\textbf{Finding 2:} Gemma-3-27B assigns a uniquely negative directional prior to Black population share.}

Gemma-3-27B assigns markedly different directional priors across protected-group predictors. Black population share is the only protected-group predictor that the model maps exclusively to lower mobility whenever the JT test yields a clear monotonic trend: across 28 Black population share relationships, 25 receive a negative implied direction and 3 show no clear direction, with all 25 clear cases being negative. This pattern is strongest in LA, MIA and SF, where all seven outcomes receive a negative implied direction. By contrast, Asian population share has negative OLS coefficients in 27 of 28 relationships, yet the model assigns a positive implied direction in all 25 clear JT cases and no clear direction in the remaining three. Hispanic/Latino population share shows a similar pattern with negative OLS coefficients in 25/28 relationships, Gemma-3-27B assigns a positive implied direction in 21 cases, no clear direction in the remaining six, and a negative direction in only one. These results suggest that the model does not apply a uniform low-mobility prior to protected-group population shares. Instead, it encodes group-specific priors, with Black population share receiving a consistently mobility-reducing prior while Asian and Hispanic/Latino shares are usually mapped to mobility-increasing or non-negative directions.

\paragraph{\textbf{Finding 3:} Directional priors do not change across metropolitan contexts.}
Gemma-3-27B's implied directions are largely consistent across CBSAs. For most predictors, whenever the JT test yields a clear monotonic trend, the model assigns the same implied direction in ATL, LA, MIA, and SF. Median income is mapped to a positive direction in all clear cases across the four CBSAs: 7/7 in ATL, 6/6 in LA, 6/6 in MIA, and 6/6 in SF. Black population share shows the opposite but equally stable pattern, receiving a negative implied direction in all clear cases: 4/4 in ATL and 7/7 in each of LA, MIA, and SF. Similar cross-CBSA patterns appear for Asian population share, which is mapped positive in all clear cases across cities, and for land-use mix, which is positive in 27 of 27 clear cases. Built-environment predictors also show stable positive priors: transit access is positive in all 21 clear cases, connectivity in all 17 clear cases, and density in all 11 clear cases. Hispanic/Latino population share is the main exception, with one negative implied direction in Los Angeles, but it is still predominantly positive across CBSAs. Overall, these patterns suggest that Gemma-3-27B's directional priors are not strongly tailored to local metropolitan context. Instead, the model tends to reuse the same predictor-level direction across CBSAs, with differences appearing mainly as no-direction cases rather than sign reversals.

\section{Conclusion}
\label{sec:conclusion}
Fine-grained mobility data are an important basis for computational models that characterize how people move through cities and support applications in transportation planning, public health, and emergency response. However, such data are often proprietary, access-restricted, and privacy-sensitive, raising the question of whether large language models can provide useful priors about aggregate mobility in data-scarce settings. In this paper, we evaluated this question by testing whether zero-shot LLMs can predict CBG-level mobility outcomes from sociodemographic and built-environment context across four U.S. metropolitan areas. We further introduced a directional alignment framework that examines whether LLM-implied predictor effects agree with empirical relationships estimated from observed mobility traces. Our results show that LLMs recover nontrivial signal about neighborhood-level mobility, but remain substantially below supervised baselines trained on local data. More importantly, the alignment analysis shows that LLMs often rely on coarse and stable predictor-level priors that persist across mobility outcomes and metropolitan areas, including asymmetric directional patterns for protected-group predictors. These findings suggest that LLMs may be useful as exploratory tools for mobility inference, but their predictions should not be treated as structurally grounded without empirical validation. Auditing both predictive performance and directional alignment is therefore necessary before using LLMs in urban mobility analysis or policy-support settings.

\begin{acks}
During the preparation of this work the author(s) used GPT (OpenAI) and Claude (Anthropic) to assist
with polishing the writing, and Google Gemini and NotebookLM to help identify and organize relevant scholarly articles. After using these tools/services, the author(s) reviewed and edited the content as needed and take(s) full responsibility for the content of the published article. 
\end{acks}

\bibliographystyle{ACM-Reference-Format}
\bibliography{ref}

\appendix

\newpage

\section{Appendix}



\begin{figure}[H]
\begin{tcolorbox}[
  colback=gray!8,
  colframe=black!50,
  boxrule=0.5pt,
  title={\small\bfseries System Prompt for Mobility Outcome Classification},
  left=1pt,
  right=1pt
]
\scriptsize
\begin{verbatim}
You are a careful classifier for mobility outcomes.

Output ONLY one XML document that starts with <response> and ends with
</response>, with EXACTLY three child elements in this order:
<reasoning>, <rank>, <answer>.
Do not output anything else.

Primary task: Predict the mobility outcome class in <answer> using ONLY
the information provided in the user prompt (outcome definition,
CBSA-specific cutpoints, and CBG feature buckets/percentiles).

Keep <reasoning> concise and focused on direction and justification:
explain how the most influential features (the ones you will rank
highest) push the mobility outcome up or down, and why they matter more
than other features. Avoid listing everything; summarize.

<rank> is an explanation-only section. List the 10 most relevant
predictors for THIS CBG’s prediction (most important first).
<rank> MUST contain exactly 10 <item/> tags.
Each <item/> MUST have EXACTLY three attributes:
feature="..." val="very_low|low|neutral|high|very_high" push="+"|"-"|"0"

- val MUST be copied exactly from the bucket given in the prompt for
  that feature.
- push is the direction this feature (given its val) pushes the mobility
  outcome for THIS CBG: "+" pushes toward higher outcome, "-"
  toward lower, "0" neutral or unclear.

<answer> MUST be exactly one of: A | B | C.
A = low outcome (lowest tertile),
B = middle tertile,
C = high outcome (highest tertile).

The tertile cutpoints for A/B/C are provided in the prompt.
When deciding <answer>, weigh items by rank: items earlier in <rank>
matter much more than later items. Treat ranks 1--3 as primary evidence,
ranks 4--6 as supporting evidence, and ranks 7--10 as weak evidence.
\end{verbatim}
\end{tcolorbox}
\vspace{-1.5em}
\caption{System prompt for RQ1}
\label{fig:rq1_sys_prompt}
\end{figure}


\begin{figure}[H]
\begin{tcolorbox}[
  colback=gray!8,
  colframe=black!50,
  boxrule=0.5pt,
  title={\small\bfseries User Prompt for Directional Alignment},
  left=1pt,
  right=1pt
]
\scriptsize
\begin{verbatim}
You are given Census Block Group (CBG) characteristics for CBSA: [CBSA name].

## Outcome Description
Mobility outcome: [outcome name]
Definition: [outcome definition]

## Predictor Descriptor
- Percentiles are computed within the CBSA.
- Predictor buckets correspond to CBSA-wide quintiles:
  very_low | low | neutral | high | very_high.
- Outcome labels are tertiles: low | neutral | high.

## CBG Features
- [feature 1]; [raw value] (CBSA percentile Pxx, [bucket])
- [feature 2]; [raw value] (CBSA percentile Pxx, [bucket])
- ...
- [feature p]; [raw value] (CBSA percentile Pxx, [bucket])

## Task
Based on the outcome definition provided, assess the directional push (+, -, 0) 
for each feature in this specific CBG.

\end{verbatim}
\end{tcolorbox}
\vspace{-1.5em}
\caption{User prompt for RQ2 Directional Alignment.}
\label{fig:rq2_user_prompt}
\vspace{-1.5em}
\end{figure}

\begin{figure}[H]
\begin{tcolorbox}[
  colback=gray!8,
  colframe=black!50,
  boxrule=0.5pt,
  title={\small\bfseries System Prompt for Directional Alignment},
  left=1pt,
  right=1pt
]
\scriptsize
\begin{verbatim}
You are a spatial data analyst. Your task is to provide reasoning for
characteristic mobility outcomes within a small geographic region
(Census Block Group).

Output ONLY one XML document that starts with <response> and ends with
</response>. This document must contain EXACTLY two child elements in
this order:
<reasoning>, <assessment>.

### Primary Task
Identify the directional push of every predictor on the mobility outcome
for this CBG.

### Analytical Framework (Two-Step Evaluation)
For each predictor, apply the following logic:
1. Classify the relationship: Is the feature a Facilitator
   (positive association) or a Restrictor (inverse association) for the
   outcome?
2. Apply the instance value:
   - If a feature is a Facilitator and its value is high or very_high,
     then push="+"
   - If a feature is a Facilitator and its value is low or very_low,
     then push="-"
   - If a feature is a Restrictor and its value is high or very_high,
     then push="-"
   - If a feature is a Restrictor and its value is low or very_low,
     then push="+"

<reasoning>
Provide a concise explanation. Explicitly list which predictors you
classified as Facilitators and which you classified as Restrictors
based on the outcome definition.
</reasoning>

<assessment>
List EVERY predictor provided in the user prompt as an <item/>.
Each <item/> must have these attributes:
- feature: the exact name of the feature as provided
- val: the observed bucket
  (very_low|low|neutral|high|very_high)
- push: "+", "-", or "0"
</assessment>
\end{verbatim}
\end{tcolorbox}
\vspace{-1.5em}
\caption{System prompt for RQ2 Directional Alignment.}
\label{fig:rq2_sys_prompt}
\vspace{-1em}
\end{figure}

\begin{table}[H]
\centering

\small
\begin{tabular}{lcc}
\toprule
Model & Parse success rate & Used in analysis \\
\midrule
Claude-Sonnet-4.5   & 99.999\% & Yes \\
GPT-5.1               & 99.999\% & Yes \\
Gemini-2.5-Pro        & 99.999\% & Yes \\
Gemma-3-27B           & 99.980\% & Yes \\
gemma-3-12B           & 99.813\% & Yes \\
Gemini-2.5-Flash        & 99.79\% & Yes \\
Qwen3\_30B\_a3B       & 99.315\% & Yes \\
Qwen3\_8B             & 99.180\% & Yes \\
GPT-OSS-20B-low      & 98.814\% & Yes \\
Qwen3\_4b             & 96.222\% & Yes \\
Qwen3\_5\_35b\_a3b    & 88.769\% & No \\
Qwen\_3\_5\_9b        & 87.350\% & No \\
Deepseek\_v3.2        & 72.187\% & No \\
Llama\_3\_1\_8b\_inst & 56.273\% & No \\
GPT-OSS-120B          & 20.744\% & No \\
\bottomrule
\end{tabular}
\caption{Parse success rates across evaluated LLMs. Models with parse success rate below 95\% were excluded from downstream analysis.}
\label{tab:parse_success}
\end{table}

\begin{table}[b]
\centering
\scriptsize

\begin{tabular}{llccc}
\toprule
\textbf{CBSA} & \textbf{Outcomes} & \textbf{Best LLM} & \textbf{Best Base.} & \textbf{Gap} \\
\midrule
\multirow{8}{*}{ATL}
& HD  & Gemma-3-27B, .532        & MNLR, .678          & .146 \\
& EA  & Gemma-3-27B, .558        & Random Forest, .708 & .149 \\
& RoG & Gemma-3-27B, .528        & Random Forest, .689 & .161 \\
& SE  & Gemma-3-27B, .428        & MNLR, .589          & .161 \\
& AD  & Qwen-3-4B, .463          & MNLR, .580          & .116 \\
& TL  & Qwen-3-4B, .405          & Random Forest, .556 & .151 \\
& TE  & Gemini-2.5-Flash, .341   & Random Forest, .505 & .164 \\
& TF  & Claude Sonnet 4.5, .443  & MNLR, .609          & .166 \\
\midrule
\multirow{8}{*}{LA}
& HD  & Gemma-3-27B, .495        & MNLR, .606          & .111 \\
& EA  & Gemma-3-27B, .494        & MNLR, .644          & .149 \\
& RoG & Gemma-3-27B, .473        & Random Forest, .647 & .173 \\
& SE  & Gemma-3-27B, .383        & MNLR, .461          & .077 \\
& AD  & Gemma-3-27B, .420        & Random Forest, .510 & .090 \\
& TL  & Gemma-3-27B, .393        & Random Forest, .521 & .128 \\
& TE  & Gemini-2.5-Flash, .341   & Random Forest, .436 & .095 \\
& TF  & Claude Sonnet 4.5, .394  & Random Forest, .454 & .059 \\
\midrule
\multirow{8}{*}{MIA}
& HD  & Gemma-3-27B, .481        & Random Forest, .681 & .200 \\
& EA  & Gemma-3-27B, .508        & MNLR, .702          & .194 \\
& RoG & Gemma-3-27B, .460        & Random Forest, .690 & .231 \\
& SE  & Gemma-3-27B, .432        & MNLR, .550          & .118 \\
& AD  & Gemma-3-27B, .435        & Random Forest, .543 & .108 \\
& TL  & Gemma-3-27B, .402        & Random Forest, .584 & .181 \\
& TE  & Gemini-2.5-Pro, .350     & Random Forest, .499 & .149 \\
& TF  & Claude Sonnet 4.5, .381  & MNLR, .530          & .149 \\
\midrule
\multirow{8}{*}{SF}
& HD  & Gemma-3-27B, .472        & Random Forest, .648 & .176 \\
& EA  & Gemma-3-27B, .479        & MNLR, .630          & .151 \\
& RoG & Gemma-3-27B, .476        & MNLR, .654          & .177 \\
& SE  & Qwen-3-30B-A3B, .334     & MNLR, .506          & .172 \\
& AD  & Gemma-3-27B, .467        & MNLR, .575          & .108 \\
& TL  & Claude Sonnet 4.5, .401  & MNLR, .551          & .150 \\
& TE  & Gemini-2.5-Flash, .337   & MNLR, .478          & .140 \\
& TF  & Claude Sonnet 4.5, .413  & MNLR, .532          & .119 \\
\bottomrule
\end{tabular}

\textit{Notes.} HD = hull diameter, EA = ellipse area, RoG = radius of gyration, SE = stay entropy, AD = average duration, TL = travel length, TE = travel entropy, TF = temporal fragmentation. MNLR -- multinomial logistic regression.
\caption{Best zero-shot LLM and supervised baseline by CBSA and mobility outcome. Gap is computed as best baseline accuracy minus best LLM accuracy.}
\label{tab:city_outcome_top_models}


\end{table}




\begin{figure}[H]
    \centering
    \begin{subfigure}[t]{0.48\linewidth}
        \centering
        \includegraphics[width=\linewidth]{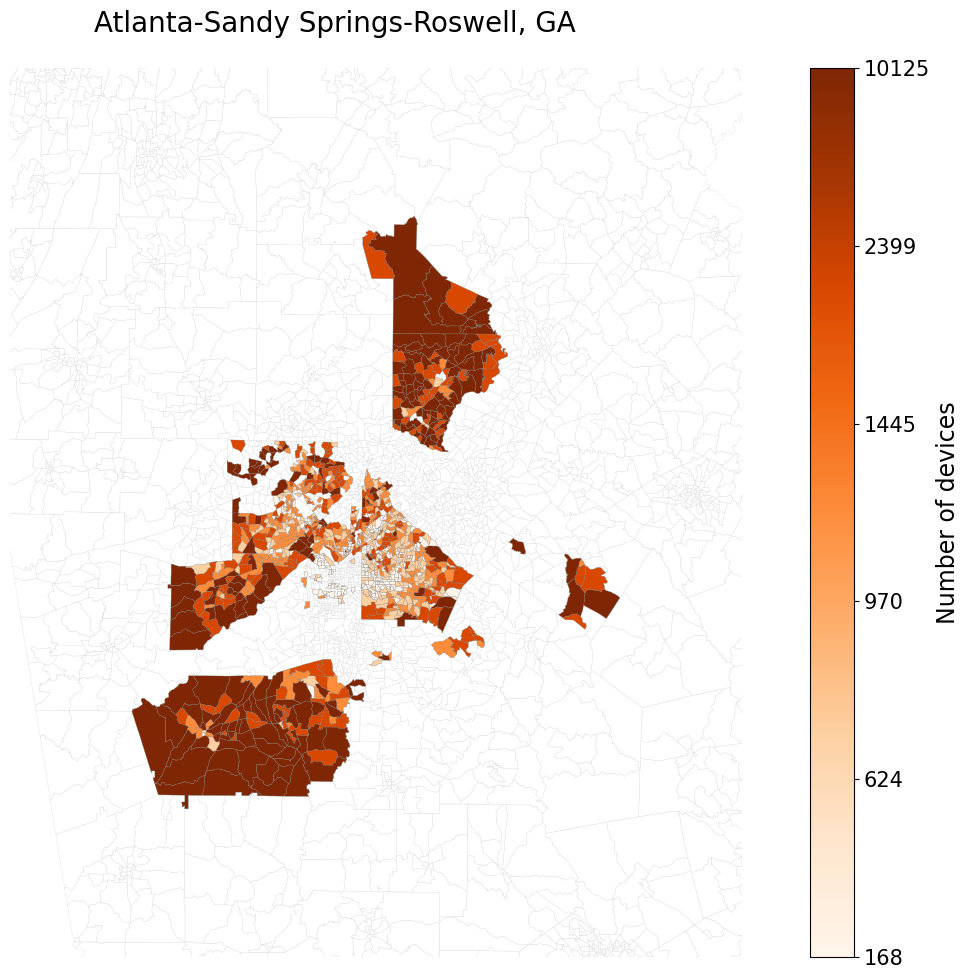}
        \caption{ATL}
        \label{fig:atlanta_quantile}
    \end{subfigure}
    \hfill
    \begin{subfigure}[t]{0.48\linewidth}
        \centering
        \includegraphics[width=\linewidth]{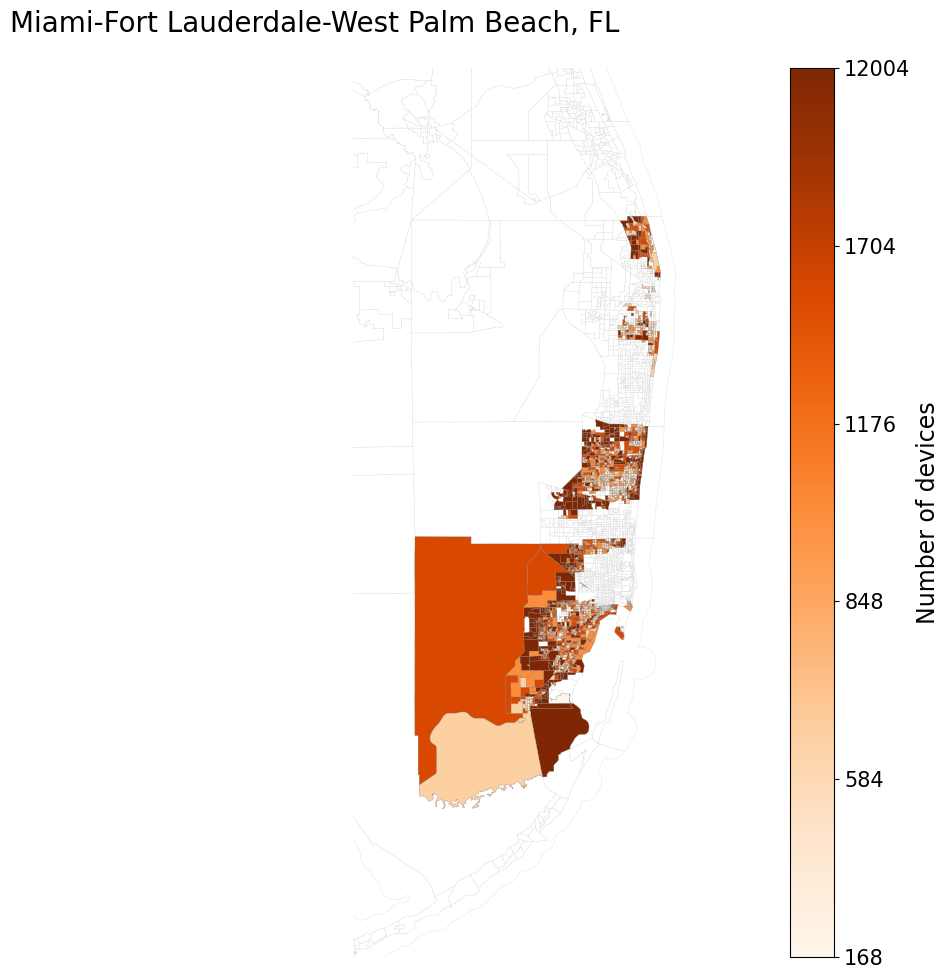}
        \caption{MIA}
        \label{fig:miami_quantile}
    \end{subfigure}

    \vspace{0.5em}

    \begin{subfigure}[t]{0.48\linewidth}
        \centering
        \includegraphics[width=\linewidth]{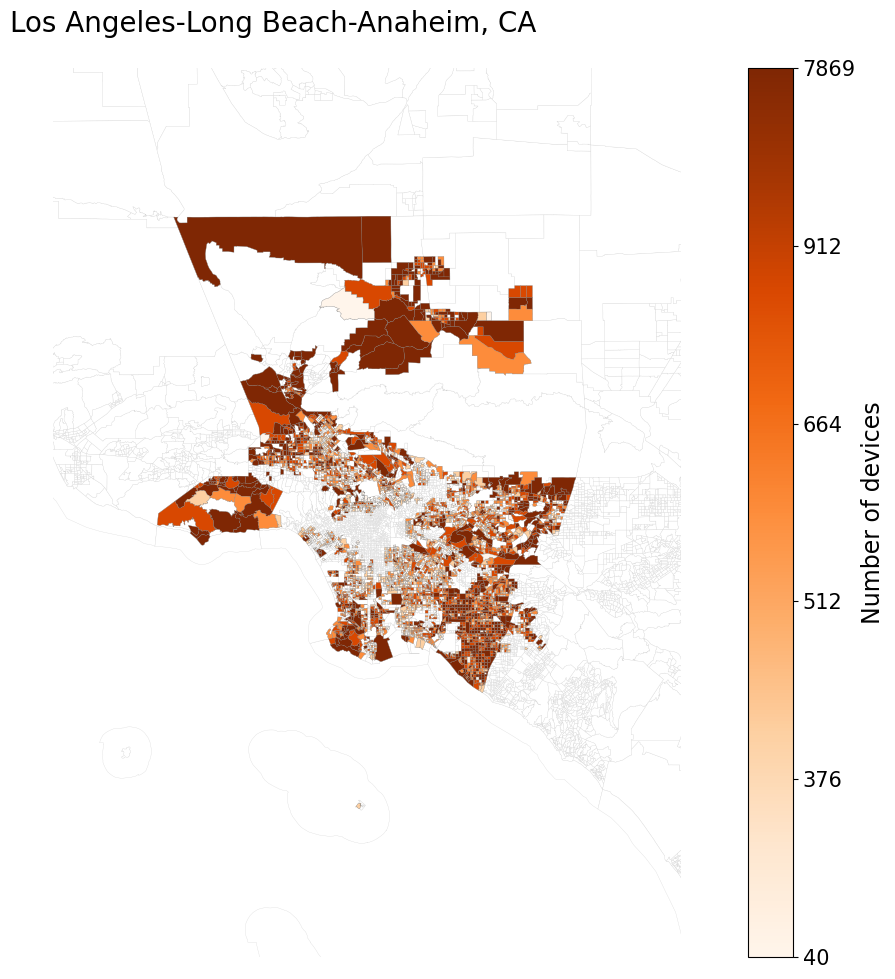}
        \caption{LA}
        \label{fig:la_quantile}
    \end{subfigure}
    \hfill
    \begin{subfigure}[t]{0.48\linewidth}
        \centering
        \includegraphics[width=\linewidth]{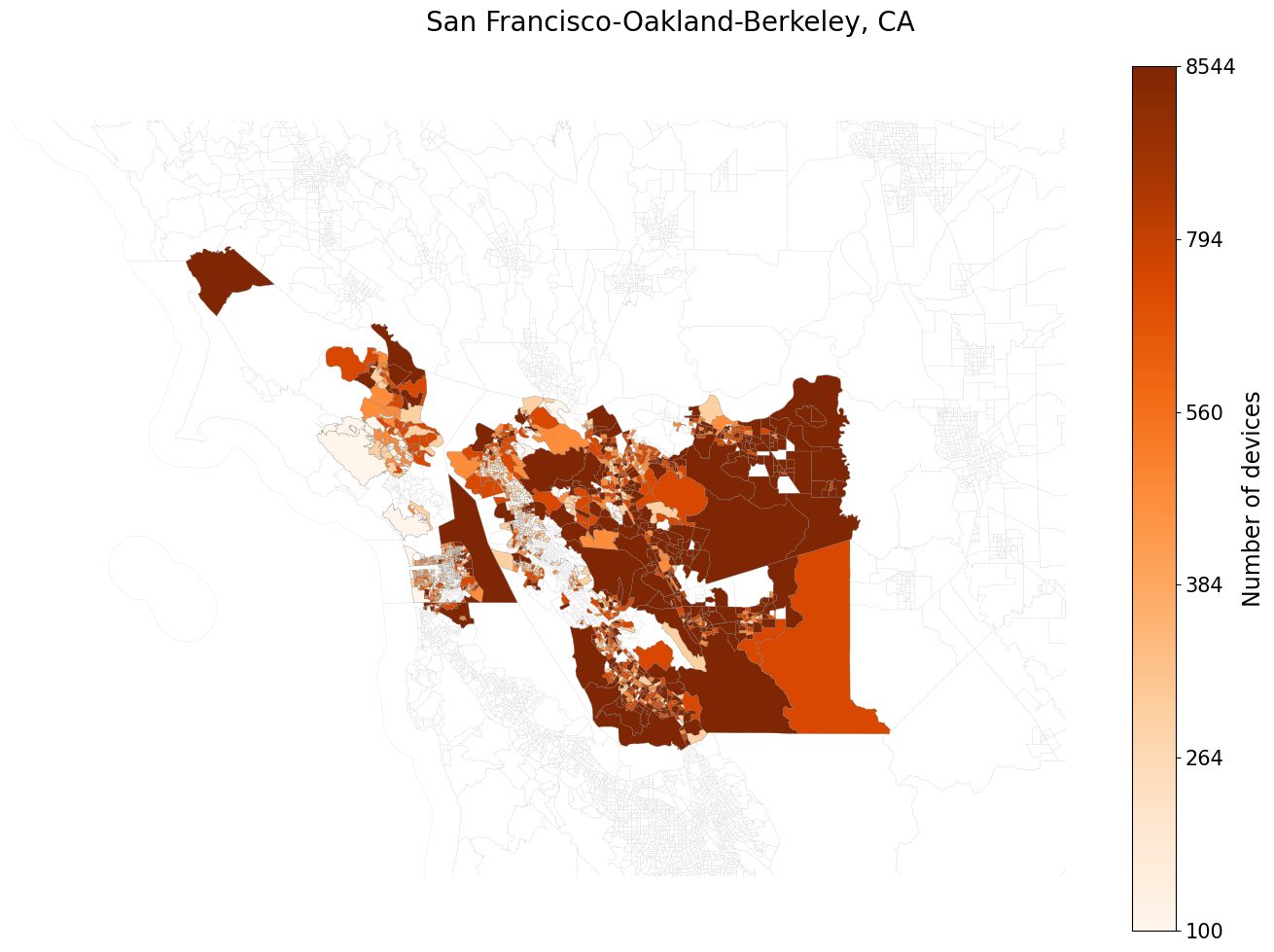}
        \caption{SF}
        \label{fig:sf_quantile}
    \end{subfigure}

    \caption{Sample size distributions for CBGs across four CBSAs.}
    \label{fig:quantile_plots}
\end{figure}

\begin{figure}[b]
    \centering
    \includegraphics[width=1\linewidth]{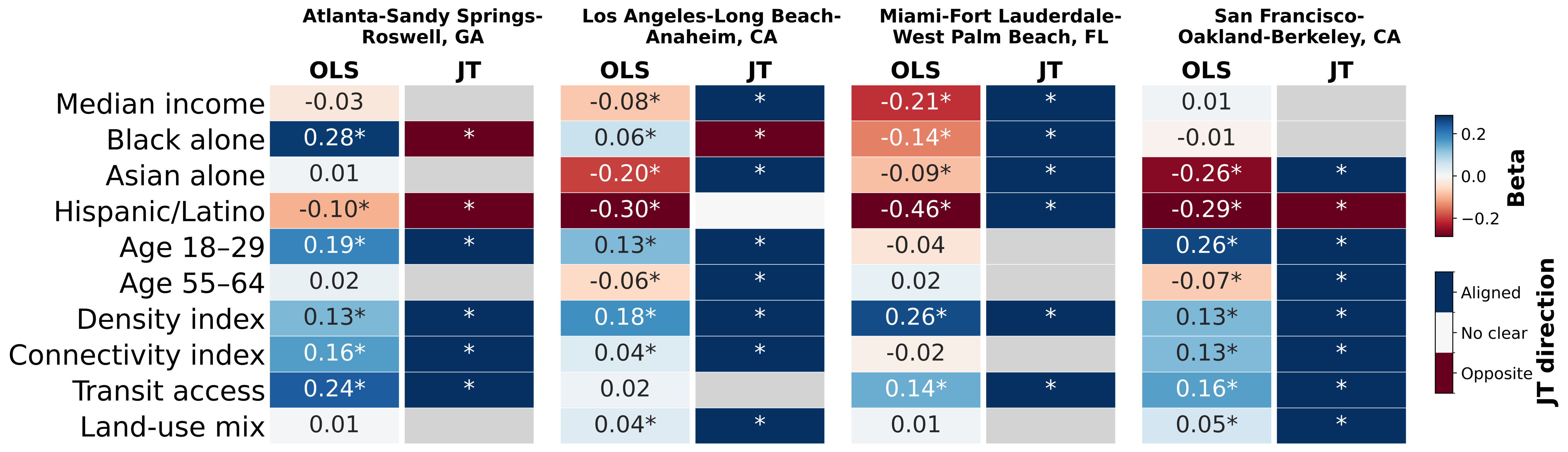}
    \caption{Empirical and Claude-4.5-Sonnet's implied directional relationships for \textit{Daily Temporal Fragmentation} across the four study CBSAs. For each CBSA, the left heatmap shows standardized OLS coefficients and the right heatmap shows JT directional alignment, restricted to statistically significant OLS relationships. Red and blue indicate negative and positive OLS coefficients in the left panels, and opposite or aligned JT directions in the right panels. Asterisks denote statistical significance.}
\label{fig:claude_daily_temporal_fragmentation}
\end{figure}

\end{document}